\documentclass[letterpaper, 10 pt, conference]{ieeeconf}
\usepackage[dvipsnames]{xcolor}
\usepackage{soul}
\usepackage{tikz}
\usepackage{multirow}
\usepackage{graphicx}
\usepackage{booktabs}
\usepackage{bm}
\usepackage{hyperref}
\usepackage{moresize}
\hypersetup{
    colorlinks=true,
    linkcolor=red,
    filecolor=magenta,      
    urlcolor=blue,
    pdftitle={darkgs},
    pdfpagemode=FullScreen,
    }
\usepackage{mathtools}
\usepackage{amsmath}
\usepackage{amssymb}
\usepackage{amsfonts}
\usepackage{subcaption}
\usepackage{caption}
\usepackage{perl_acronyms}
\let\labelindent\relax
\usepackage[inline]{enumitem}
\usepackage{siunitx}
\usepackage[
    style=ieee,
    doi=false,
    isbn=false,
    url=false,
    eprint=false,
    backend=bibtex,
    natbib=true
    ]{biblatex}
\IEEEoverridecommandlockouts                            
\title{\LARGE \bf DarkGS: Learning Neural Illumination and 3D Gaussians Relighting for Robotic Exploration in the Dark}

\author{Tianyi Zhang$^{1}$, Kaining Huang$^{1}$, Weiming Zhi$^{1}$, Matthew Johnson-Roberson$^{1}$
\thanks{$^{1}$Authors are with the Robotics Institute, School of Computer Science,
        Carnegie Mellon University,
        Pittsburgh, PA 15213, USA
        {\tt\small \{tianyiz4,kaining2,wzhi,mkj\}@andrew.cmu.edu}}%
}

\begin{document}
\maketitle
\thispagestyle{empty}
\pagestyle{empty}
\begin{abstract}
Humans have the remarkable ability to construct consistent mental models of an environment, even under limited or varying levels of illumination. We wish to endow robots with this same capability. In this paper, we tackle the challenge of constructing a photorealistic scene representation under poorly illuminated conditions and with a moving light source.
We approach the task of modeling illumination as a learning problem, and utilize the developed illumination model to aid in scene reconstruction.
We introduce an innovative framework that uses a data-driven approach, \emph{Neural Light Simulators} (NeLiS), to model and calibrate the camera-light system.
Furthermore, we present DarkGS, a method that applies NeLiS to create a relightable 3D Gaussian scene model capable of real-time, photorealistic rendering from novel viewpoints. We show the applicability and robustness of our proposed simulator and system in a variety of real-world environments. Code released at \url{https://tyz1030.github.io/proj/darkgs.html}
\end{abstract}

\section{INTRODUCTION}
\label{sec:introduction}
\par
\begin{figure}[t]
  \centering 
  \begin{subfigure}[b]{0.92\linewidth} 
    \includegraphics[width=\textwidth]{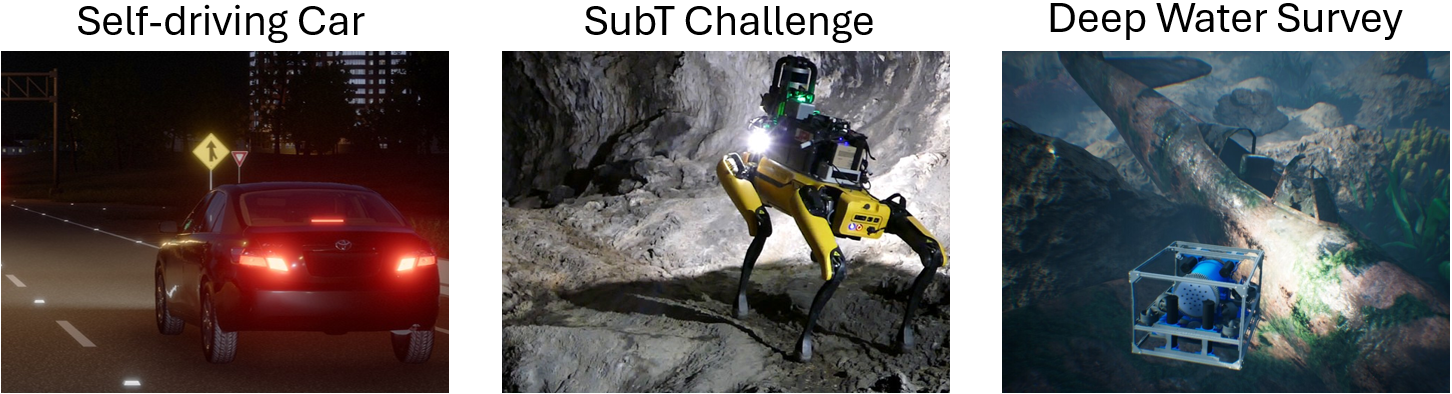}
    \caption{Robots in the dark with onboard light sources}
    \label{intro:teasera}
  \end{subfigure}
  \vfill 
  \vspace{0.5cm}
  \begin{subfigure}[b]{0.92\linewidth} 
    \includegraphics[width=\textwidth]{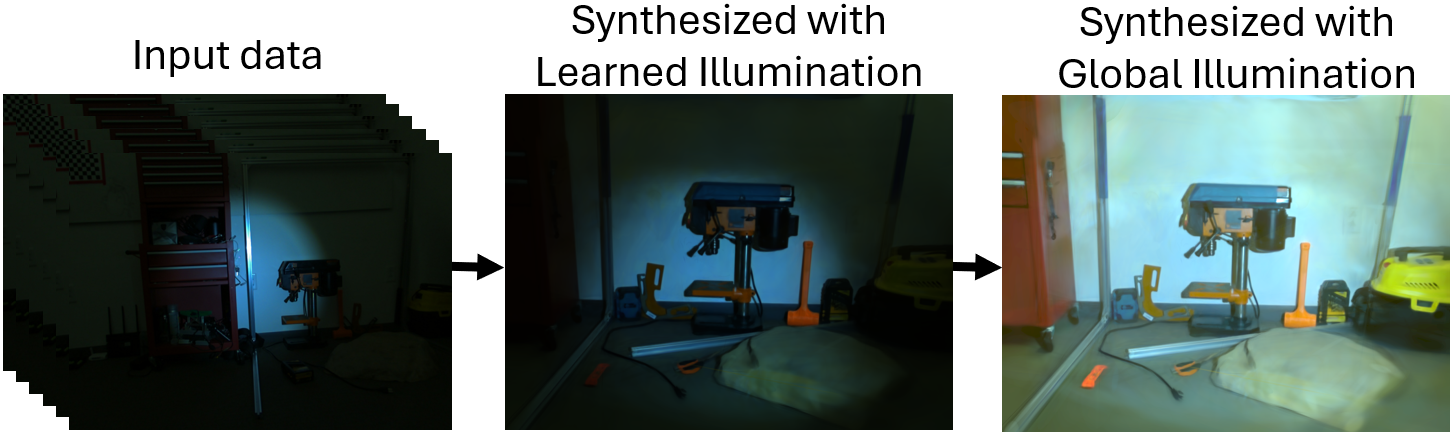}
    \caption{Our work build 3D Gaussians and relight the scene in dark}
    \label{intro:teaserb}
  \end{subfigure}
  \caption{Robotic imaging systems working in the dark consist of cameras and light sources. Examples as shown in (a): Carla Simulator~\cite{carla}, Team CoStar in SubT Challenge~\cite{costarnebula} and HoloOcean underwater robot simulator~\cite{holoocean}. In this work, we propose a pipeline that calibrates the camera-light system which helps photorealistic scene reconstruction and relighting from images collected in the dark.}
  \label{fig:teaser}
\end{figure}

 Robots and autonomous vehicles have been routinely deployed in poorly illuminated environments for critical missions and tasks such as exploration, inspection, transportation, search and rescue, etc. (see Fig.~\ref{intro:teasera}). Imaging systems consisting of one or multiple RGB cameras and light sources are often equipped on the robot to illuminate and sense the surrounding environment. The streamed image sequence can be further used in downstream tasks, e.g. navigation, mapping, and visualization, to boost the robot autonomy and human understanding of the environment.

\par Scene reconstruction, or the capability to create accurate internal representations of the environment, is vital for robots operating in unknown environments.
Previous vision-only approaches largely rely on identifying common feature points over a set of multiple-view images, and then minimizing a reprojection error~\cite{buildrome}. Such procedures like \ac{SfM} or \ac{SLAM} also estimate the camera poses of the images.
Based on these camera poses,  \ac{NeRF}~\cite{mildenhall2020nerf} is capable of achieving photorealistic scene reconstruction by optimizing a photometric loss between the representation and the images. However, while achieving huge success in the graphics community, the transition of objective function from reprojection error to photometric loss has raised a new challenge to the robotics community: Can we still build consistent scene representations with a moving light source on the robot platform?

\par In this work, we identify \emph{illumination-inconsistency} as the main challenge in building a photorealistic scene representation from images taken with a moving light source. That is, as the robot operates in the environment, the effect of moving light source results in images captured of the same region appearing visually different. Handcrafted feature descriptors that serve as backbone in \ac{SfM} and \ac{SLAM}~\cite{sift, murAcceptedTRO2015, ORBSLAM3_TRO} are designed to be invariant to changes in pixel intensities. Such changes also have limited effects on the optimization pipeline, which uses reprojection errors as objective functions.
Hence, the classic \ac{SfM} methods are generally robust against \emph{illumination-inconsistency}. However, on the way towards photorealistic scene reconstruction and switching to optimization of the photometric loss, we find that state-of-the-art \ac{NeRF} methods~\cite{mildenhall2020nerf, kerbl3Dgaussians} fail on the images taken in the dark with a moving light source. Although variants such as RawNeRF~\cite{mildenhall2022rawnerf} and Relightable 3DG~\cite{R3DG2023} claim to be able to handle dark images in RAW format and model the reflection properties of the scene, they still fail to handle the varying illumination.

Concretely, the problem tackled in this paper is as such: Given a sequence of images taken in poorly illuminated environments, with one major light source moving with one camera as a rigid body, reconstruct the scene by minimizing photometric loss and achieve photorealistic novel view image synthesis (Fig.~\ref{intro:teaserb}). Our contributions are as follows:
\begin{enumerate}
\item A pipeline that consists of light source modeling, camera-light calibration, building 3D Gaussians and scene relighting from illumination-inconsistent images. 

\item \ac{NeLiS}, a data-driven and physically interpretable illumination model and software for light source modeling and calibration.

\item \ac{DarkGS}, a variant of the \ac{3DGS} model that builds photorealistic scene representations under poorly illuminated conditions and relights the scene with global illumination, based on COLMAP~\cite{colmap} and \ac{NeLiS} results.
\end{enumerate}

\section{RELATED WORK}
\label{sec:relatedwork}
\subsection{Extroceptive Sensor Calibration on Robots}

\begin{figure*}[t]%
\centering
\includegraphics[width=0.90\linewidth]{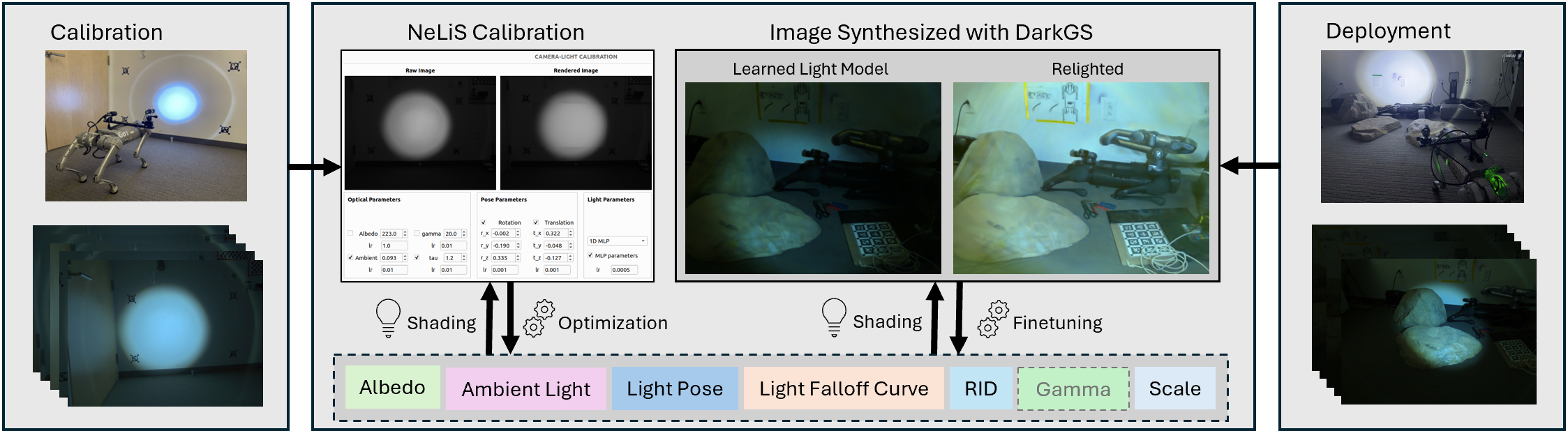}
    \caption{Our proposed workflow: Images for camera-light calibration are first collected at a calibration target. With \ac{NeLiS}, we manually initialize the parameters and then optimize the light model. The model can then be used to build \ac{DarkGS}, present the scene with learned or relighted illumination. (Learning  Gamma tone mapping is supported by \ac{NeLiS} but not further discussed in this paper.)}
    \label{method:banner}
\end{figure*}

Autonomous robots are usually equipped with perceptual sensors such as cameras and LiDARs. Taking camera calibration as an example, a calibration target with AprilTag~\cite{apriltag} or checkerboard pattern is often used. With observations of the target from different perspectives, the focal length, center of projection, distortion coefficient, and pose of the camera can be estimated~\cite{rehder2016extendingkalibr}.
Similar target-based approaches have been used to calibrate LiDARs~\cite{beltran2022camlidar}, radars~\cite{radarcameracalib} and acoustic sensors~\cite{sonarcalib} for downstream sensor fusion tasks~\cite{song2024lirafusion}.
Analogous to sensor calibrations mentioned above, in this paper, we propose to use a target consisting of AprilTags~\cite{apriltag} and blank space for calibrating the light in a camera-light system, including estimating the transformation between camera's and light's coordinate system, \ac{RID} and light fall-off curve of the light source.

\subsection{Light Calibration}
Existing methods use various kinds of customized calibration targets:~\cite{santo2018light}\cite{santo2020light} propose to use a target of AprilTags and pins to infer the position of point light source from the shadows of the pins. Alternatively,  ~\cite{fastled2021}~\cite{Ma:23} propose to use a Lambertian sphere and estimate the light source parameters by learning to reconstruct the sphere. 
Furthermore, \cite{ShadowBasedLight} shows that the location of the light source can also be recovered from the shadow of a sphere. With a Lambertian plane and customized markers, ~\cite{Ma:19} shows that it is possible to calibrate a point light source with a single image. \cite{song2020light} shows the most relevance to our work, which calibrates the pose of a light source and the metric scale given the RID curve. However, the above work either assumes an oversimplified model, e.g. a point light source with inverse square light falloff, in a pure dark environment, or that the \ac{RID} is given, which makes the method less applicable. In this work, we use a minimum calibration target design that consists of only AprilTags~\cite{apriltag} and a Lambertian plane similar to~\cite{song2020light}. With \ac{NeLiS} we show that modeling and learning \ac{RID}, light falloff and ambient light together with camera-light pose can improve the quality of image reconstruction and make our calibration method more feasible for real-world robot deployments.

\subsection{NeRF and 3D Gaussian Splatting}
Based on the success of \ac{SfM}~\cite{buildrome}, \ac{NeRF} learns to represent scenes using a continuous function, e.g. a \ac{MLP}, and can achieve photorealistic novel-view image synthesis~\cite{chen2024survey}. Popular variations of \ac{NeRF} such as Mip-NeRF\cite{Barron2021MipNeRFAM} and Instant-NGP~\cite{mueller2022instant} have improved \ac{NeRF} by rendering accuracy and speed. Recent innovation introduced as \ac{3DGS}~\cite{kerbl3Dgaussians} has revolutionized this field by replacing the neural network backbone with spatial Gaussians, allowing rendering speed of 100+ fps while maintaining the accuracy and differentiability of the system.
For our robotic setup, the problem of varying illumination is attempted by~\cite{neuralref, neuralsea} which model the physical property of the scene, e.g. roughness, albedo, and normal, instead of the radiance as a constant. However, these works oversimplify the light model as a Lambertian point light source, co-centered with the camera, making it not applicable on many real robotic platforms.
Gaussianshader~\cite{jiang2023gaussianshader} and Relightable 3DG~\cite{R3DG2023} introduce physical properties into the \ac{3DGS} framework but model illumination as a constant that does not change frame-by-frame. According to our experiments, none of the above mentioned methods handles the illumination-inconsistancy issue on a real-robot imaging system. In our proposed \ac{DarkGS}, by modeling the physical property of the scene and taking advantage of \ac{NeLiS}, we can not only build a consistent \ac{3DGS} from poorly illuminated images, but also relight the scene with global illumination.

\subsection{Photometric Stereo}
Photometric stereo refers to the estimation of the shape of an object with multiple images taken from a single viewpoint with different lighting orientations~\cite{photometricstereo}. Early approaches confine this problem to known light source and Lambertian surfaces with uniform albedo. Further work has allowed spatially varying albedo~\cite{chakrabarti2016single} and unknown light~\cite{photounknownlight}. However, our robotic setup differs from this problem by the fact that our camera and light source always move together as a rigid body. And we choose to tackle this problem within the framework of \ac{3DGS}~\cite{kerbl3Dgaussians} which allows us to synthesize novel views with photorealistic quality.

\section{METHODOLOGY}
\label{sec:methodology}
Our proposed workflow for building \ac{NeLiS} and \ac{DarkGS} is shown in Fig.~\ref{method:banner}. We build \ac{NeLiS} by calibrating the camera-light system at a planar calibration target.
By accounting for the calibration image sequence and hand measurement of the light pose, an in-the-loop human operator can initialize the light parameters using our intuitive GUI.
Since the calibration target can be localized by AprilTags~\cite{apriltag}, a shading model can render the image with the given parameters to approximate the calibration image. Then, we run the optimization to minimizes the photometric error of the calibration image sequence and rendered image. After obtaining a calibrated \ac{NeLiS}, the robot can be deployed in a dark environment, using the collected images to build a \ac{DarkGS}, finetune the model, and relight the scene.

In this section, we will discuss the details of the \ac{NeLiS} model in~\ref{sec:shadingmodelcali}, \ac{DarkGS} in~\ref{sec:shadingmodel3dgs} and \ac{NeLiS} frontend software and workflow in ~\ref{gui}. 

\subsection{Shading Model for NeLiS}\label{sec:shadingmodelcali}
\begin{figure}[t]%
\centering
    \includegraphics[width=0.94\linewidth]{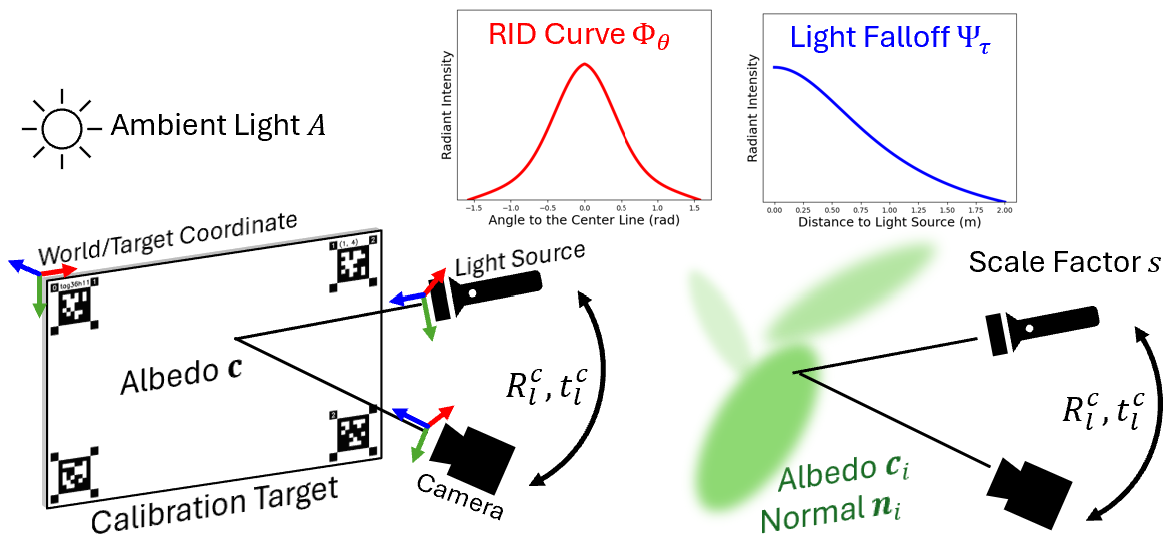}
    \caption{Our shading model: (Left) In \ac{NeLiS}, camera poses are localized by AprilTags on the calibration target. The pose of light $R_l^c$ and $t_l^c$, albedo $\mathbf{c}$ of the calibration target, ambient light $A$, RID $\Phi_\theta$, and light falloff function $\Psi_\tau$ will be learned. (Right) In building \ac{DarkGS}, each Gaussian $g_i$ is modeled with an albedo $\mathbf{c}_i$ and normal $\mathbf{n}_i$. The ambient light $A$ and scale $s$ will also be optimized in this process.}
    \label{method:calibmodel}
\end{figure}
The calibration data is taken by capturing photos at a calibration target from different views while the light source moves with the camera as a rigid body.
The calibration target is a white plane with four AprilTags positioned as four corners of a rectangle (as shown in Fig.~\ref{method:calibmodel}). We attach the origin of the world coordinate to the top-left corner of the calibration target.
We assume that the camera is precalibrated with distortions removed. One of the key problems for \ac{NeLiS} to solve is estimating the relative pose of light to the camera $R_l^c \in {SO(3)}$ and $t_l^c\in \mathbb{R}^3$.

Given the true size of the calibration target, we can apply \ac{PnP}~\cite{pnp} to extract 
$R_c^w\in SO(3)$ and $t_c^w\in \mathbb{R}^3$ which transform points from the camera coordinate to the world coordinate.
The position of the light source is then given by $\mathbf{o}_l = R_c^w t_l^c+t_c^w\in \mathbb{R}^3$. Since we align the $z$ axis of the light coordinate with the centerline of the light, the direction of the centerline can be denoted by $\boldsymbol{\omega}_{l} = R_c^wR_l^c[0,0,1]^\top \in \mathbb{R}^3$. Both $\mathbf{o}_l$ and $\boldsymbol{\omega}_l$
are in the world coordinate frame.

\par We only use the area bounded by 4 AprilTags as the \ac{ROI} to do calibrations. We assume that this area is a Lambertian plane and has the same normal $\mathbf{n}\in\mathbb{R}^3 $ and diffusive albedo $\mathbf{c}\in\mathbb{R}^{\lambda}$ anywhere on the plane ($\lambda = 1$ for grayscale images and $\lambda = 3$ for RGB). For each pixel in the \ac{ROI}, we find the intersection of the corresponding camera ray and the target plane in the world coordinate system, denoted by $\mathbf{x}\in \mathbb{R}^3$.
To infer the incident radiance at $\mathbf{x}$, one needs to model the \ac{RID}, light falloff function, and ambient light.

\subsubsection{RID} \ac{RID} is commonly modeled as a function of the angle between the centerline $\boldsymbol{\omega}_{l}$ of the light and light ray $\boldsymbol{\omega}_\mathbf{x} = \mathbf{x}-\mathbf{o}_l$.
Previous work assume that \ac{RID} is given~\cite{song2020light, Ma:19, Ma:23}. However, this assumption may not hold for in-the-wild robot deployment. Instead, we remove this dependency by learning a neural \ac{RID} from calibration data:
\begin{equation}
    \Phi_\theta(\mathbf{x}) 
    = \text{MLP}_\theta(\cos^{-1}(\frac{\boldsymbol{\omega}_\mathbf{x}}{\lVert \boldsymbol{\omega}_\mathbf{x} \rVert_2}\cdot \boldsymbol{\omega}_{l}))
\end{equation}
here $\theta$ denotes the parameters of the MLP.
\subsubsection{Light Falloff Curve} Inverse square law is widely used to model light falloff, based on the assumption of a point light source. However, when objects are closer to the light source, the inverse square law starts to fail. We choose to model the light falloff in the form of a Lorentzian function of the distance $\lVert \boldsymbol{\omega}_\mathbf{x} \rVert_2$, as suggested by~\cite{ryer1997light}:
\begin{equation}
    \Psi_\tau(\mathbf{x}) = \frac{1}{\tau+\lVert \boldsymbol{\omega}_\mathbf{x} \rVert_2^2}
\end{equation}
Instead of estimating $\tau$ from hand measurement of the light surface~\cite{ryer1997light}, we designate it to be a learnable parameter.

\subsubsection{Ambient Light}
Imaging and lighting problems are often studied in a dark room to better simplify and constrain the model by removing ambient light. However, a perfectly dark space for calibration might not be accessible for real-world robot deployment. Instead, we model ambient light as a learnable parameter $A$. The incident radiance at the point $\mathbf{x}$ can thereby be modeled as:

\begin{equation}
\label{eq:incid}
    I_\mathbf{x} = \Psi_\tau(\mathbf{x}) \Phi_\theta(\mathbf{x})+A
\end{equation}

\subsubsection{BRDF} Although there are \ac{BRDF} models capable of handling complex reflection effects, they exceed the scope of this paper. We opt for the Lambertian reflection model, eliminating the need to optimize any parameters in our \ac{BRDF}. We use $f_r(\boldsymbol{\omega}_\mathbf{x}, \mathbf{n}, \mathbf{c}) = \text{max}(\boldsymbol{\omega}_\mathbf{x}\cdot \mathbf{n},0)\mathbf{c}$ as our \ac{BRDF} (Lambert cosine law included in \ac{BRDF} for simplicity, following convention of NeRV~\cite{nerv2021}
), giving the rendering equation:
\begin{equation}\label{eq:rendercalib} 
\hat{L}_\mathbf{x} = I_\mathbf{x} f_r(\boldsymbol{\omega}_\mathbf{x}, \mathbf{n}, \mathbf{c})
\end{equation}
With captured pixel intensity $L_\mathbf{x}$, we use L1 loss and formulate the \ac{NeLiS} optimization problem as:
\begin{equation}
    \min_{\theta, A, \tau, R_l^c, t_l^c, {\mathbf{c}}} \sum_{\mathbf{x}\in{\text{ROI}}} \|L_\mathbf{x}-\hat{L}_\mathbf{x}\|_1
\end{equation}

\subsection{Building DarkGS}\label{sec:shadingmodel3dgs}
Within the framework of \ac{3DGS}, we model the scene with a point cloud of Gaussians $\mathbf{G}$ (as shown in Fig.~\ref{method:calibmodel} right). Each Gaussian ${g}_i$ in the point cloud encompasses attributes including position $\mathbf{p}_i$, covariance $\Sigma_i$, opacity $\alpha_i$, albedo $\mathbf{c}_i$ and normal $\mathbf{n}_i$, that ${g}_i = \{\mathbf{p}_i, \Sigma_i, \alpha_i, \mathbf{c}_i, \mathbf{n}_i\}\in \mathbf{G}$.

Given $\mathbf{p}_i$, the incident radiance $I_i$ can be calculated by Eq.~\ref{eq:incid}. With $\mathcal{N}$ ordered points for pixel $(u,v)$, the rendering equation then becomes:

\begin{equation}\label{eq:rendergs}    
\hat{L}_{u, v} =\sum_{i\in\mathcal{N}} I_i f_r(\boldsymbol{\omega}_i, \mathbf{n}_i, \mathbf{c}_i) \alpha_i \prod_j^{i-1} (1-\alpha_j)
\end{equation}

\subsubsection{Scale Recovery} The framework of \ac{3DGS} and its variants are often based on monocular \ac{SfM} solutions such as COLMAP~\cite{buildrome}. While calibration using AprilTags recovers poses in true metric scale with \ac{NeLiS}, monocular \ac{SfM} only gives up-to-scale poses for building 3D Gaussians. Here, we introduce a scaling factor $s>0$ as a learnable parameter so that we can obtain the positions with scale $\mathbf{p}_i^\prime =s\mathbf{p}_i$.
With captured pixel intensity $L_{u, v}$, the 3D Gaussian can be built by solving the follow optimization problem:
\begin{equation}
\label{eq:render3dgs}
    \min_{A, \mathbf{G}, s} \sum\|L_{u, v}-\hat{L}_{u, v}\|_1
\end{equation}

\subsubsection{Training Warm-Up} In some cases, a large discrepancy between the initial scale and the true scale often leads to divergence and local minimum at the beginning of optimization. We therefore propose a warm-up strategy to tackle this problem. The overall idea behind this warm-up strategy is that we overwrite our light pose to be co-centered and parallel with the camera. In this co-centered configuration, the baseline between camera and light will be zero, so that the scale difference between the baseline and the point cloud will have a minor effect on shading at coarse level. Then, we gradually recover the pose to the calibration result over the first $k$ iterations. During this process, Eq.~\ref{eq:render3dgs} is being solved and $s$ will be optimized. We define a warm-up factor that grows with iterations, so that in the $m^{th}$ iteration, this factor is $\frac{m}{k}$. Here, we denote Lie exponential map and Lie logarithm map as follows: 
\begin{equation}
\text{exp}(\cdot):\mathbb{R}^3\rightarrow SO(3),\ \ \text{log}(\cdot):SO(3)\rightarrow \mathbb{R}^3
\end{equation}

so that in warm-up stage we replace $R_l^c$ and $t_l^c$ with:
\begin{equation}
  \hat{R}_l^c = \text{exp}(\frac{m}{k}\text{log}(R_l^c)),\ \ \hat{t}_l^c = \frac{m}{k}t_l^c
\end{equation}

\subsubsection{Relighting}
Once the \ac{DarkGS} model is built, we can relight the scene by replacing the components in Eq.~\ref{eq:incid}, i.e. $\Psi_\tau$, $\Phi_\theta$ and $A$, by carefully designed values and functions. For example, replacing the MLP function in $\Phi_\theta$ with a constant can create Lambertian illumination without a strong pattern; replacing the light falloff function $\Psi_\tau$ with a constant can simulate the illumination from a parallel light source.

\subsection{NeLiS frontend: Human-in-the-loop calibration}\label{gui}
Although front-end software is not the main contribution of this paper, it provides an interactive interface for \ac{NeLiS} visualization, manual initialization of parameters, and hyperparameter tuning, as partially shown in Fig.~\ref{method:banner} and elaborated on our \href{https://github.com/tyz1030/neuralight.git}{github}. We find that the \ac{NeLiS} model is prone to local minimum, especially when learning an MLP and camera-light pose from scratch at the same time. In addition, optimizing MLP and albedo together creates an under-constrained problem, often leading to numerical instability. With our GUI, we first initialize the camera-light pose with a Gaussian distribution as initial \ac{RID}, and fine-tune the pose, albedo, and other parameters. Then we train an MLP-based \ac{RID} with pose and albedo freezed, preventing divergence in concurrently learning multiple parameters. Once the model has converged, we unfreeze all the learnable parameters and optimize them together to reach the global minimum.



\section{EXPERIMENTS}
\label{sec:experiments}
\subsection{Experiments Setup}
\begin{figure}[t]%
\centering
    \includegraphics[width=0.95\linewidth]{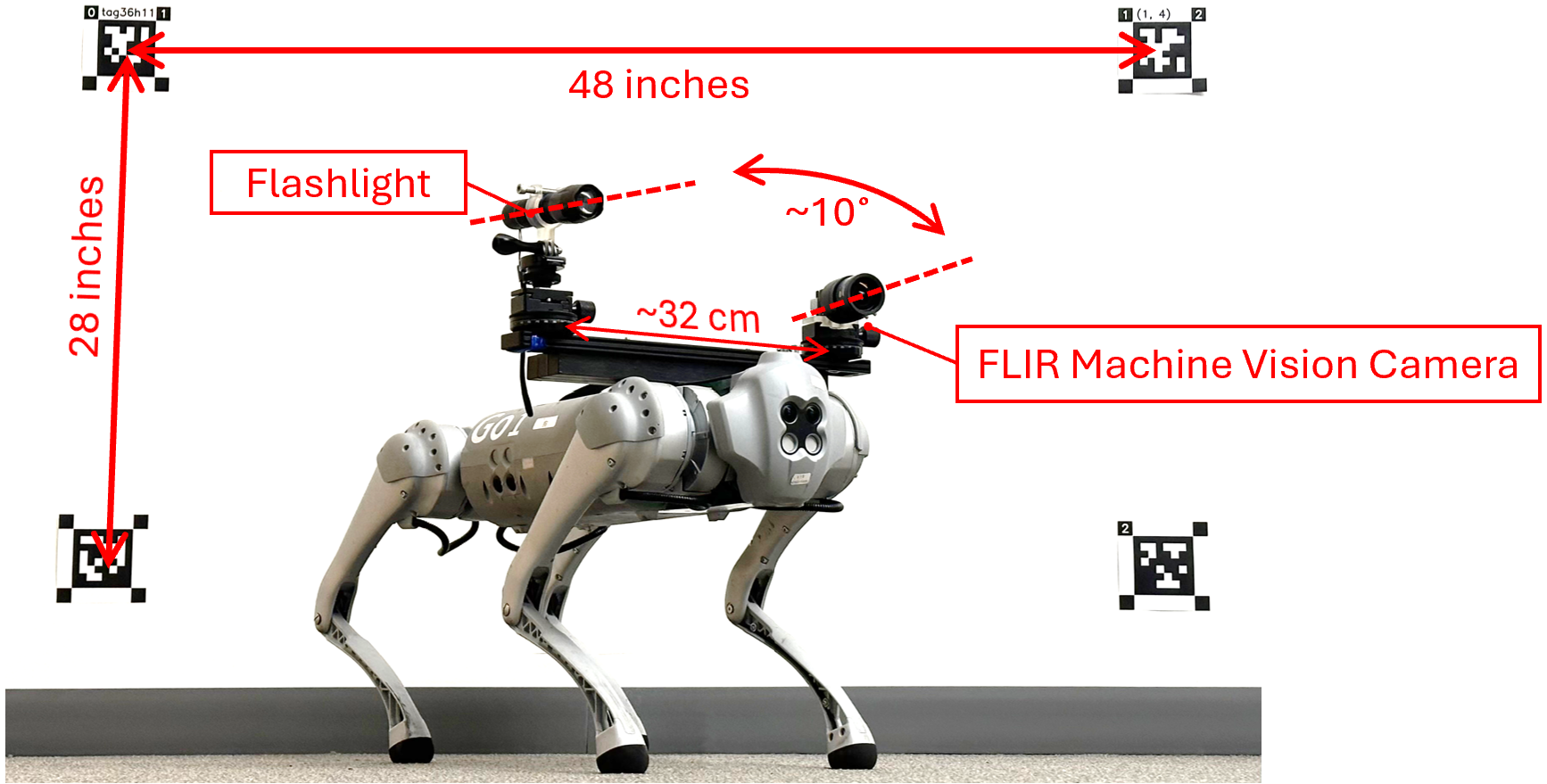}
    \caption{Our experiment setup: The imaging system is installed on a legged robot platform (Unitree GO1). We use a FLIR machine vision camera to stream the images in RAW format. The calibration target (as shown behind the dog) is a white wall with four AprilTags positioned in a rectangle.}
    \label{exp:setup}
\end{figure}
Our imaging system consists of a FLIR machine vision camera (Model Firefly S) and a light source, together mounted on a rigid structure on top of our legged robot, as shown in Fig.~\ref{exp:setup}. The baseline between the camera and the light source is approximately 32 cm, measured by hand.
Although the ray-tracing problem can be simplified by placing the light source and camera close enough and modeling them as co-centered~\cite{neuralref}, in many applications a long baseline between them is a preferred design worth studying. For example, deep-sea robots usually have a long baseline between light and camera to avoid massive light backscattering from the water.
\par We experiment with three different light sources: a flashlight, a diving light, and a flood light. To build \ac{NeLiS}, we take $\sim40$ calibration images from different ranges and perspectives for each light configuration and use $25\%$ of the images as a testing set to evaluate the \ac{NeLiS} performance. For building \ac{DarkGS}, we take $50\sim150$ images for each scene. All images are streamed in RAW format without any tone mapping or white balancing. Parameters optimized in both \ac{NeLiS} and \ac{DarkGS} use Adam optimizer~\cite{adam} with LieTorch~\cite{teed2021tangent} integration.

\subsection{Can existing 3DGS methods do the job?}

We first investigate whether existing 3DGS methods can reconstruct the scene using the images we collected with our robotic setup in the dark. We experimented with: \begin{enumerate*}
  \item vanilla \ac{3DGS}~\cite{kerbl3Dgaussians} which models the scene with constant radiance 
  \item RawNeRF~\cite{mildenhall2022rawnerf} which is developed to reconstruct scene from RAW HDR images
  \item Relightable \ac{3DGS}~\cite{R3DG2023} which models the shading and environmental light map. 
\end{enumerate*}
To make a fair comparison, we apply gamma tone mapping to images for vanilla \ac{3DGS}~\cite{kerbl3Dgaussians} and Relightable \ac{3DGS}~\cite{R3DG2023} which brightens the image and smooth the discrepancy between illuminated and under-illuminated areas. We also replace RawNeRF's backbone MLP-based renderer~\cite{mildenhall2022rawnerf} with Gaussian render~\cite{kerbl3Dgaussians}.

\begin{figure}[t]%
\centering
    \includegraphics[width=0.99\linewidth]{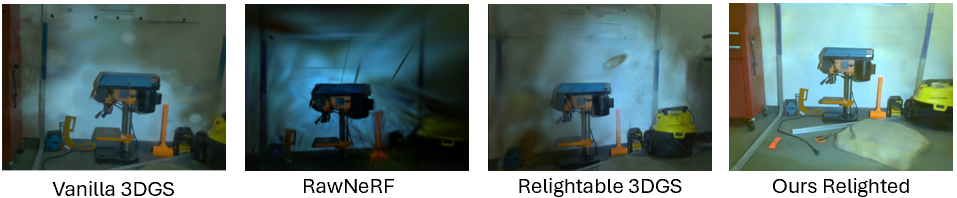}
    \caption{None of the existing methods can solve the problem: Results of Vanilla 3DGS~\cite{kerbl3Dgaussians}, RawNeRF~\cite{mildenhall2022rawnerf} Relightable 3DGS~\cite{R3DG2023} show heavy artifacts and fail to converge. The key reason for the failures is that the existing method does not model the illumination change.}
    \label{exp:vanilla}
\end{figure}

As shown in Fig.~\ref{exp:vanilla}, we find that none of the existing methods is able to build valid scene reconstructions. We see an excessive amount of artifacts in the center area of the synthesized image. The key reason is that the capability to model varying illumination is missing from existing methods. 

\begin{figure}[ht]
  \centering 
  \begin{subfigure}[b]{0.26\linewidth} 
    \includegraphics[width=\textwidth]{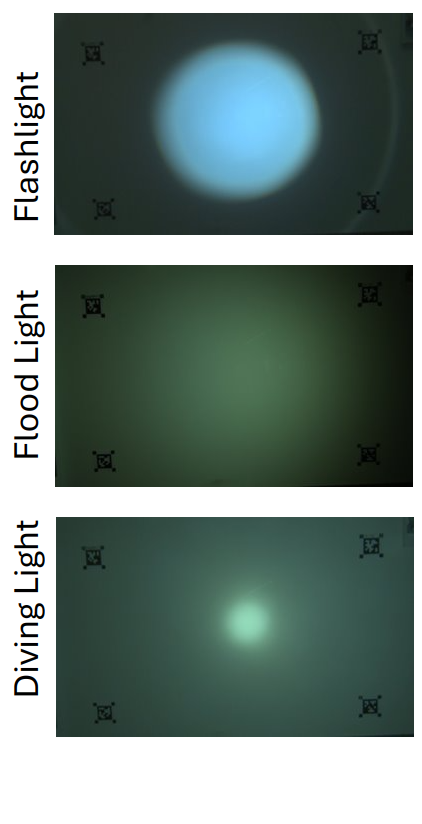}
    \caption{\ac{RID}}
    \label{exp:rid}
  \end{subfigure}
  \hspace{0.1cm} 
  \begin{subfigure}[b]{0.69\linewidth} 
    \includegraphics[width=\textwidth]{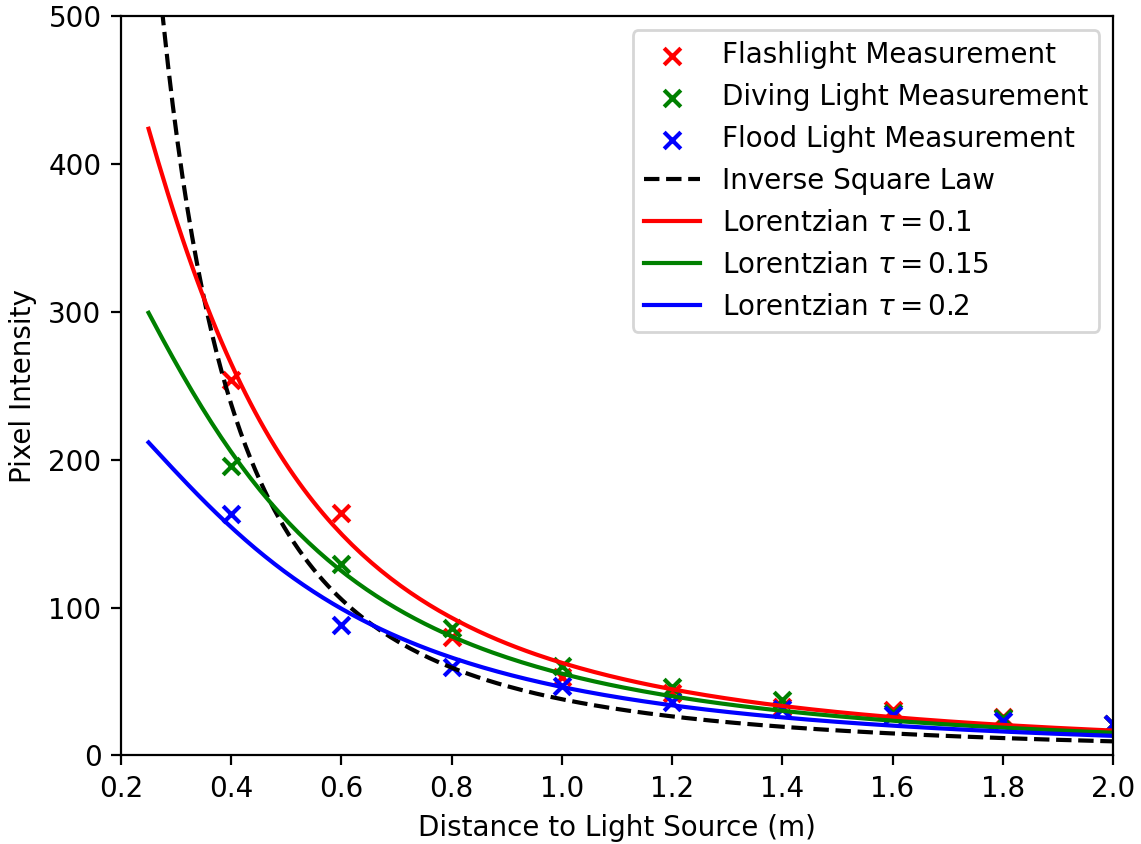}
    \caption{Light Fall-off Curve}
    \label{exp:falloff}
  \end{subfigure}
  \caption{(a) Light sources on real robots have various \ac{RID} patterns; (b) Real world measurements of light falloff show that the inverse square law is insufficient to model any of our light sources, but Lorentzian functions~\cite{ryer1997light} with learnable parameter $\tau$ fit them well.}
  \label{fig:ridandtau}
\end{figure}

\subsection{Why do we need to learn RID?}
\begin{figure}[t]%
\centering
    \includegraphics[width=0.90\linewidth]{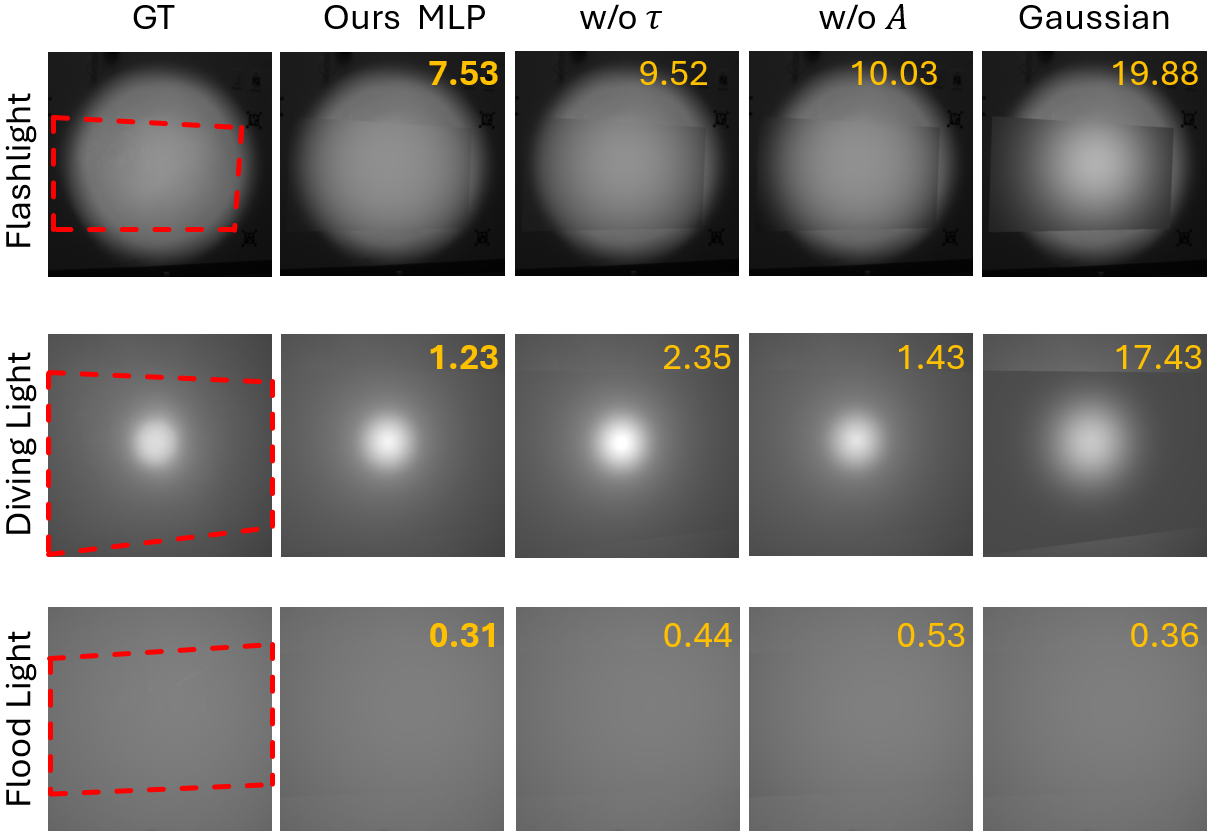}
    \caption{Ablation study: We show that our model with an MLP-based \ac{RID}, learnable light fall-off parameter $\tau$ and ambient light $A$ can effectively improve the rendering performance. \colorbox{Gray}{\textcolor{Goldenrod}{MSE}} of the testing set are highlighted in yellow. The
    \textcolor{red}{red dashed box}
bounds the regions to be rendered.}
    \label{exp:abl}
\end{figure}

As shown in Fig.~\ref{exp:rid}, different light has different \ac{RID} patterns. Existing methods often model \ac{RID} as known~\cite{song2020light}, or general functions such as the power of cosine functions~\cite{Ma:23} or a Gaussian distribution. However, for most light sources, and all light sources we used in this research, \ac{RID} is not given. Modeling them with general functions such as a Gaussian distribution will not be adaptive and expressive enough to reflect the true \ac{RID} of different light patterns for building a photorealistic renderer. The ablation study in Fig.~\ref{exp:abl} (columns 2 and 5) shows that while a Gaussian distribution fits
the \ac{RID} of flood light well, it performs much worse on the other two types of light. In comparison, our MLP-based model fits all 3 kinds of light with the best performance, showing good adaptivity to different light patterns.

\begin{figure*}[t]%
\centering
\includegraphics[width=0.98\linewidth]{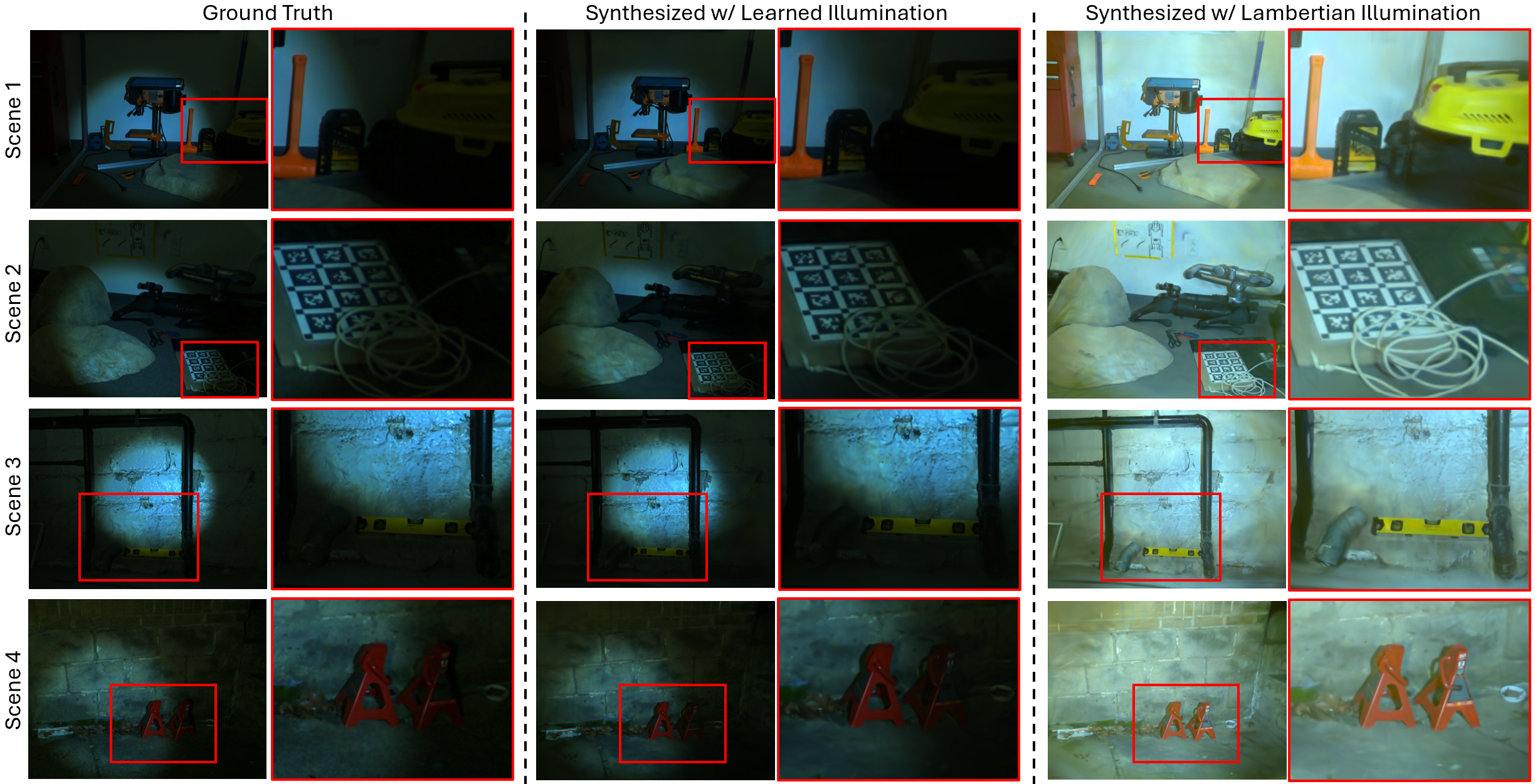}
    \caption{Visulization of results from multiple scenes: We show that with \ac{DarkGS}, we can reconstruct the scene with RAW images from robotic deployments in dark environments, and relight the scene to reveal more information that is corrupted in the RAW image due to uneven and partial illumination. Results as shown are all from the flashlight setup which is the most challenging according to our numerical results.}
    \label{exp:viz}
\end{figure*}

\begin{figure}[h]%
\centering
\includegraphics[width=0.98\linewidth]{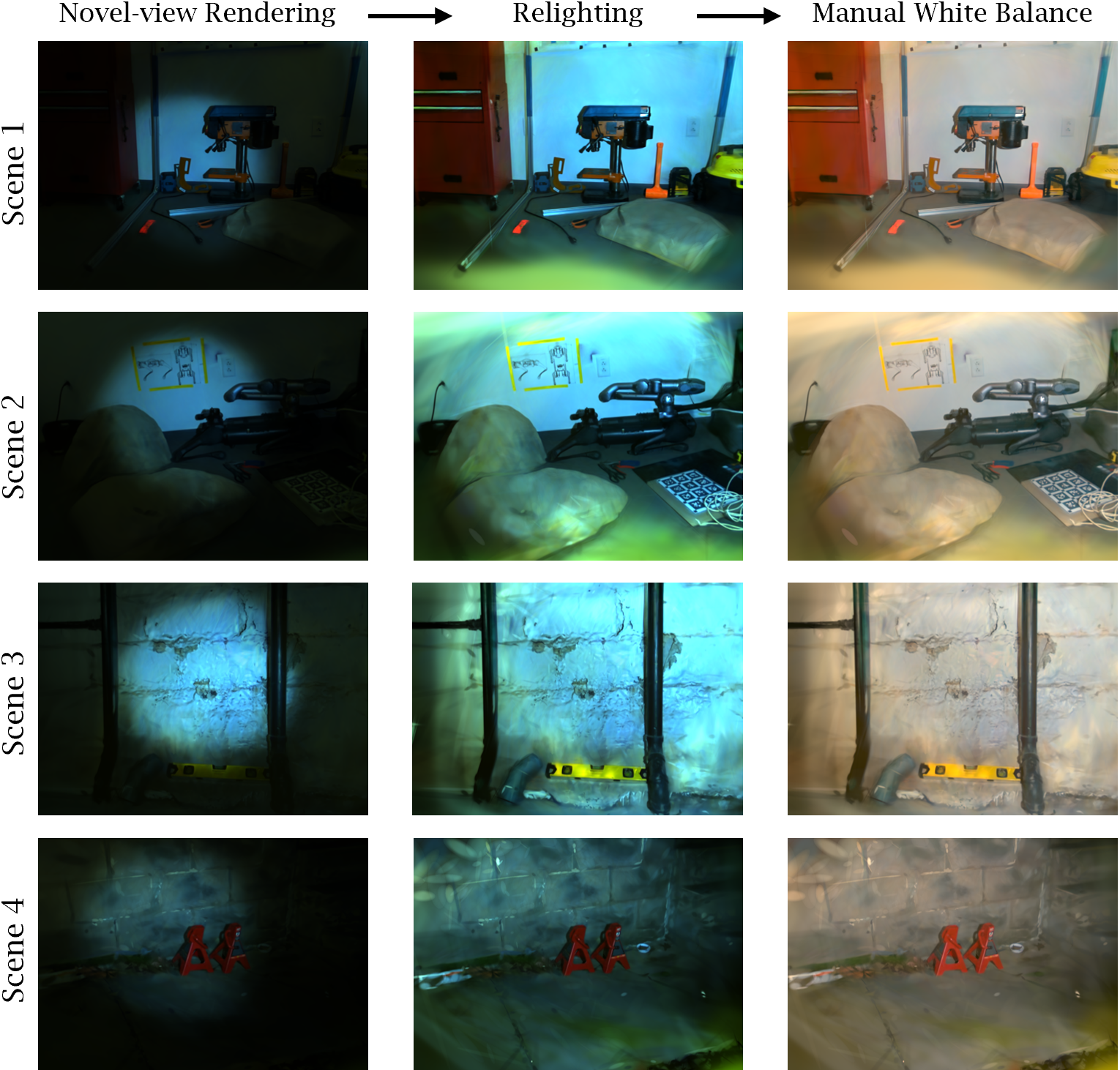}
    \caption{Novel-view rendering, relighting with virtual Lambertian illumination, and manual white balance.}
    \label{exp:novel}
\end{figure}

\subsection{Why do we need to learn light falloff?}
For simplicity, previous studies~\cite{Ma:19,Ma:23,song2020light} use an inverse square law to model the light falloff. It is not a good fit for real-world light systems. We measure the light falloff of the center point of 3 different light sources as shown in Fig.~\ref{exp:falloff}. The plot show that when the range is close, light fall-off does not follow the inverse square law as the point light source assumption starts to fail. However, with a Lorentzian model suggested by~\cite{ryer1997light}, the light falloff can be better fitted with our introduced learnable parameter $\tau$. We further show numerical results in Fig.~\ref{exp:abl} (columns 2 and 3) that by learning $\tau$, the rendering performance on testing set of all 3 kinds of lights gets improved, which implies that our light fall-off model better approximates the true light falloff.

\subsection{Do we need a perfectly dark environment?}
No. Ambient light is universal in all kinds of environment and a perfectly dark room for calibration is rare for robot deployments in the wild. Neither our calibration with \ac{NeLiS} nor deployment experiments with \ac{DarkGS} are performed in a perfectly dark environment. As shown in Fig.~\ref{exp:abl} (columns 2 and 4), we show that with our modeling of ambient light $A$, the performance of the testing set is improved on all three different lights.

\subsection{Results Visualization}
We deploy our system in various real-world environments and the results are shown in Fig.~\ref{exp:viz}. Compared with ground truth, our model is able to reconstruct the image with photorealistic quality with learned illumination. Then we replace the light source in our model with a virtual Lambertian light to create a global illumination that illuminates the entire \ac{FOV} of the camera with photorealistic rendering quality.

We further visualize the novel-view rendering results in Fig~\ref{exp:novel}. In the first column in Fig~\ref{exp:novel}, we show that the learned light effects generalize to novel views well, with visual quality comparable to training views in Fig.~\ref{exp:viz}. We relight the scene (column 2) in a way similar to Fig.~\ref{exp:viz}, and manual white balance (column 3).

\section{LIMITATIONS}
\label{sec:discussions}

\begin{figure}[ht]
  \centering 
  \begin{subfigure}[b]{0.41\linewidth} 
    \includegraphics[width=\textwidth]{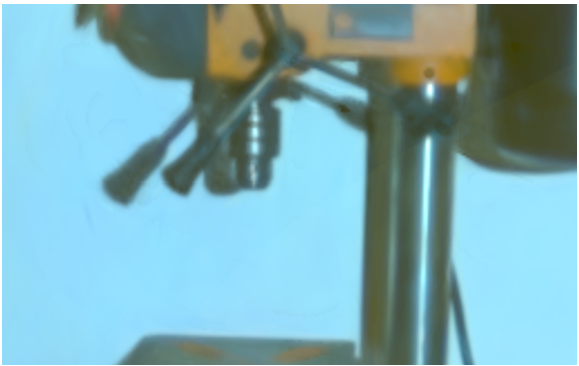}
    \caption{Shadow \& Specularity}
    \label{diss:lima}
  \end{subfigure}
  \hspace{0.2cm} 
  \begin{subfigure}[b]{0.41\linewidth} 
    \includegraphics[width=\textwidth]{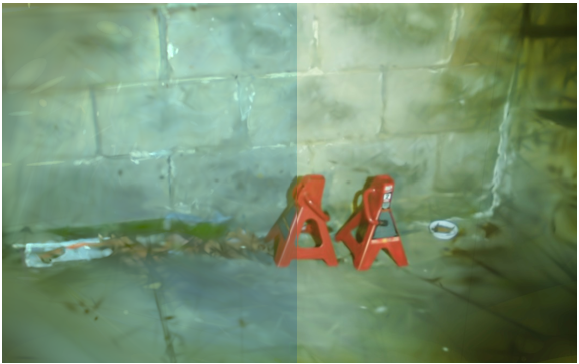}
    \caption{Auto White Balancing}
    \label{diss:limb}
  \end{subfigure}
  \caption{Limitations: (a) Our current method is not able to model shadows and specularity. (b) The relighted image is still in green-blue tone due to the nature of RAW images format and that automatic white balancing is missing from our pipeline.}
  \label{fig:lim}
\end{figure}

A limitation of our work is that our model currently does not support modeling objects with heavy shadows or specular reflection (Fig.~\ref{diss:lima}). However, recent concurrent works~\cite{bolanos2024gaussianshadow, R3DG2023} have shown the potential to integrate them into the \ac{3DGS} framework. It is also obvious that our relighted image is in green-blue tone (Fig.~\ref{diss:limb}). This is because \begin{enumerate*}
    \item We are sticking with the RAW image format throughout the pipeline, so images appear green as the RAW image nature
    \item Our LED light has a high color temperature (cool color) so that the relighted image appears blue.
\end{enumerate*}

\section{CONCLUSIONS AND FUTURE WORK}
\label{sec:conclusions}
\par This work aims to solve the problem of building 3D Gaussians and scene relighting from images taken by a moving camera-light system. Our proposed pipeline consists of \ac{NeLiS}, a camera-light simulation and calibration model, and \ac{DarkGS} which build's photolistic representation for scenes in the dark. The results show that our proposed pipeline can build relightable Gaussians from images taken by the robot platform deployed in the dark, while other current approaches are not able to do the job. Ablations show that components in our proposed model effectively improve the performance of our shading model, so we can learn \ac{RID} with an MLP for arbitrary light pattern, better approximate the light-falloff curve, and allow us to calibrate and deploy the system in environments that are not perfectly dark. These efforts make our model more applicable in real-world robot deployments. Future work includes modeling shadows and non-Lambertian objects, and bringing automatic white balance and tone mapping into the loop.

\section*{ACKNOWLEDGEMENTS}
\label{sec:acknowledgements}
This work is supported by NOAA NA22OAR0110624.



\renewcommand{\bibfont}{\normalfont\small}
\printbibliography
\end{document}